%% file: main.tex
\documentclass[10pt,twocolumn,letterpaper]{article}
\usepackage[utf8]{inputenc}
\usepackage{cvpr}              
\usepackage{graphicx}
\input{preamble}

\definecolor{cvprblue}{rgb}{0.21,0.49,0.74}
\usepackage[pagebackref,breaklinks,colorlinks,citecolor=cvprblue]{hyperref}

\def\confName{ECCV}
\def\confYear{2024}

\title{OCC-MLLM-Alpha:Empowering Multi-modal Large Language Model for the Understanding of Occluded Objects with Self-Supervised Test-Time Learning.}

\author{
Shuxin Yang\\
Amazon Robotics\\
shuxin.y.97@gmail.com\\
\and
Xinhan Di\\
Giant Network AI Lab\\
dixinhan@ztgame.com
}
\begin{document}
\maketitle
\input{sec/0_abstract}
\input{sec/1_intro}
\input{sec/3_method}
\input{sec/4_dataset}
\input{sec/5_Experiment}
{
    \small
    \bibliographystyle{ieeenat_fullname}
    \bibliography{main}
}


\end{document}

%% file: preamble.tex
\usepackage[dvipsnames]{xcolor}


%% file: sec/0_abstract.tex
\begin{abstract}
There is a gap in the understanding of occluded objects in existing large-scale visual language multi-modal models. Current state-of-the-art multi-modal models fail to provide satisfactory results in describing occluded objects through universal visual encoders and supervised learning strategies. Therefore, we introduce a multi-modal large language framework and corresponding self-supervised learning strategy with support of 3D generation. We start our experiments comparing with the state-of-the-art models in the evaluation of a large-scale dataset SOMVideo \cite{Zhang_2024_CVPR}. The initial results demonstrate the improvement of $16.92\%$ in comparison with the state-of-the-art VLM models.
\end{abstract}

%% file: sec/1_intro.tex
\section{Introduction}
\label{sec:intro}
The latest multi-modal dialogue models \cite{chen2023shikra,gong2023multimodal,yang2023mm,chen2023lion,lin2024moe,liu2024visual,alayrac2022flamingo,jin2022expectation,gao2023llama,li2023blip,wu2023visual,zhu2023minigpt4,ye2023mplug}, such as Mini-Gemini \cite{li2024mgm} and GPT-4o \cite{gpt4o} showed that despite significant progress, their description of large-scale language models for occluded objects remains unsatisfactory.

\begin{figure*}[htbp]
    \centering
    \includegraphics[width=\textwidth]{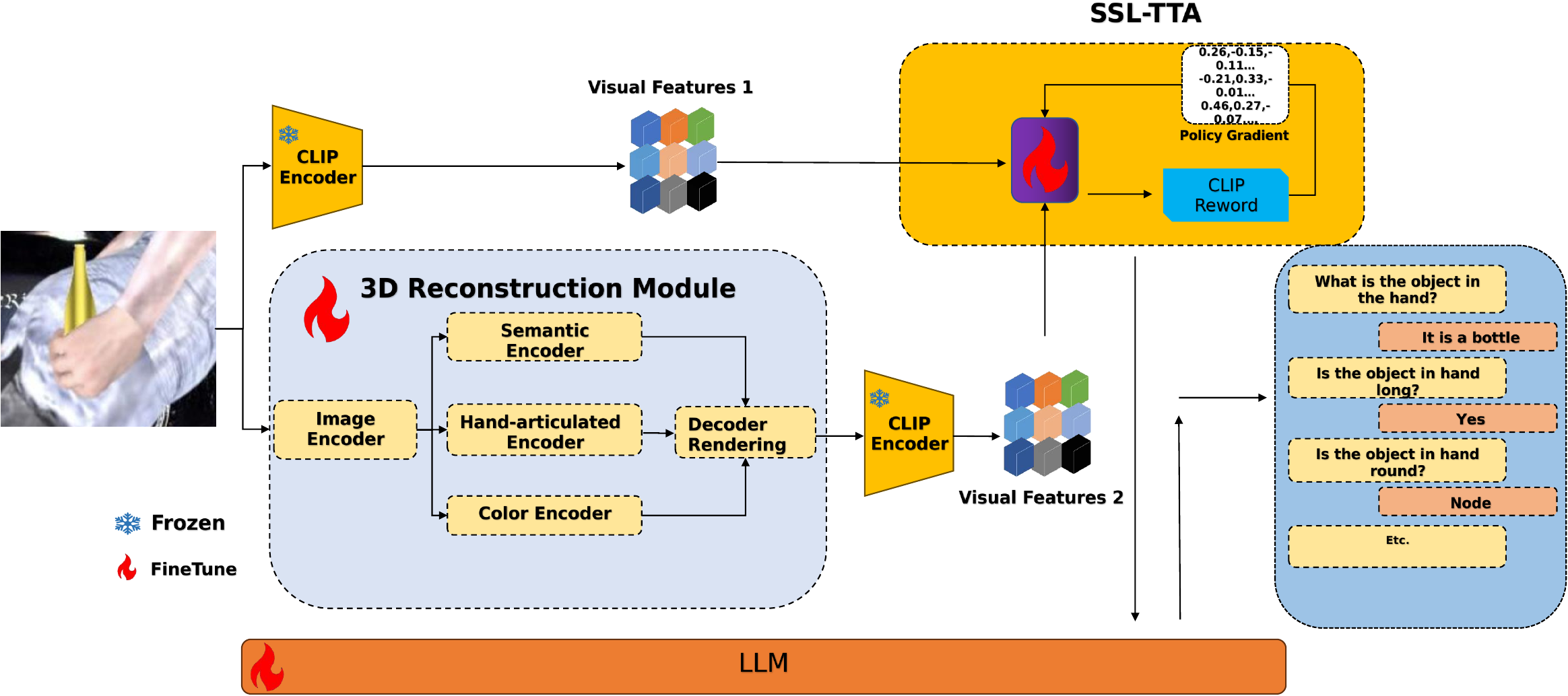}
    \caption{Overview of the Proposed Multi-Modal Vision-Language Model for the Occluded Objects with Self-Supervised Test-Time Learning.}
    \label{fig:fig1}
\end{figure*}

Therefore, we propose OCC-MLLM-Alpha, a visual language model (shown in Figure \ref{fig:fig1}) designed to understand occluded objects in image conversations. To achieve this goal, we developed a visual encoder module consisting of the common CLIP model \cite{radford2021learning} and the proposed 3D model \cite{chen2023gsdf}. Additionally, a self-supervised test-time learning strategy with the support of 3D generation is proposed.

%% file: sec/3_method.tex
\section{Method}
First, we formulate the generative process of the proposed MLLM, named Occlusion-Aware Multimodal Large Language Model (OCC-MLLM-Alpha), for occlusion-aware descriptions of objects at hand. Second, we introduce the formulation details of each proposed OCC-MLLM-Alpha module. Third, the proposed occlusion loss is calculated, and an occlusion-aware training strategy for large multi-modal language models is introduced. Fourth, a self-supervised test-time training strategy is designed to facilitate the understanding of occluded objects. We represent the generation process of the proposed OCC-MLLM-Alpha into three parts: input formulation, model forwarding, and decoding.

\subsection{Formulation of OCC-MLLM-Alpha Generation}
\subsubsection{Input Formulation} 
The input of the proposed OCC-MLLM-Alpha consists of images and text. Setting aside specific architectural differences, OCC-MLLM-Alpha generally applies a visual encoder module to extract visual tokens from raw images and uses a cross-modal mapping module to map these tokens to text space as the input of LLM. The mapped visual tokens are used as part of the LLM input along with the text input. The visual tokens are represented as $\mathbf{x}^v=\left\{x_0, x_{1}, \ldots, x_{N-1}\right\}$. $N$ represents the length of the visual token, which is a fixed number in most cases. Similarly, the input text is tokenized and expressed as  $\mathbf{x}^p=\left\{x_N, x_{N+1}, \ldots, x_{M+N-1}\right\}$. The image and text tokens are then concatenated as the final input $\left\{x_i\right\}_{t=0}^{T-1}$ where $T=N+M$.

\subsubsection{Model Forward}
First, OCC-MLLM-Alpha is trained in an auto-regressive manner using causal attention masks, where each token predicts the next token based on the previous token, formally:
\begin{equation}
\begin{aligned}
& \mathbf{h}=\operatorname{F_{MLLM^{Occ}}}\left(\mathbf{x}_i\right) \\
& \mathbf{h}=\left\{h_0, h_1, \ldots, h_{T-1}\right\}
\end{aligned}
\end{equation}
where $\mathbf{h}$ represents the output hidden states of the last layer of the
$\operatorname{F_{MLLM^{Occ}}}$.

Second, the hidden state $h$ is projected by applying the vocabulary head $\mathcal{H}$ via $\operatorname{F_{MLLM^{Occ}}}$. Get the predicted logits (probability) of the next token, and the calculation is as follows:

\begin{equation}
p\left(x_t \mid x_{<t}\right)=\operatorname{SoftMax}\left[\mathcal{H}\left(h_t\right)\right]_{x_t}, \quad x_t \in \mathcal{X},
\end{equation}

where $x_{<t}$ is represented to simplify the sequence $\left\{x_i\right\}_{i=0}^{t-1}$ and $\mathcal{X}$ is represented as the whole vocabulary set.

\subsubsection{Decoding} After applying logits $p\left(x_t \mid x_{<t}\right)$, several decoding strategies have been deployed, including greedy decoding, Beam Search \cite{boulanger2013audio}, etc. The decoded tokens are concatenated to the last one of the original input text for the next generation round until the end of the generation. The proposed OCC-MLLM-Alpha applies a beam search strategy \cite{boulanger2013audio}, which is a decoding strategy based on cumulative scores.

\subsection{Dual Visual Encoder Module}
In the forwarding process of the proposed OCC-MLLM-Alpha, we designed a new visual encoder module, which consists of two visual encoders. The first visual encoder is the common CLIP \cite{radford2021learning}, which is used to extract the visual embedding (token) $x_v$ from the RGB input $\mathbf{x}_{\mathrm{v1}}$ without a specific occlusion representation. The second visual encoder is used to provide a representation of the occluded object visual embedding(token) $\mathbf{x}_{\mathrm{v2}}$. Then, the combined representation is calculated as follows:
\begin{equation}
\mathbf{x}^{v}=\alpha \cdot \mathbf{x}^{v1}+(1-\alpha) \cdot \mathbf{x}^{v2}
\label{eq3}
\end{equation}

\begin{figure}
    \centering
    \includegraphics[width=0.35\textwidth]{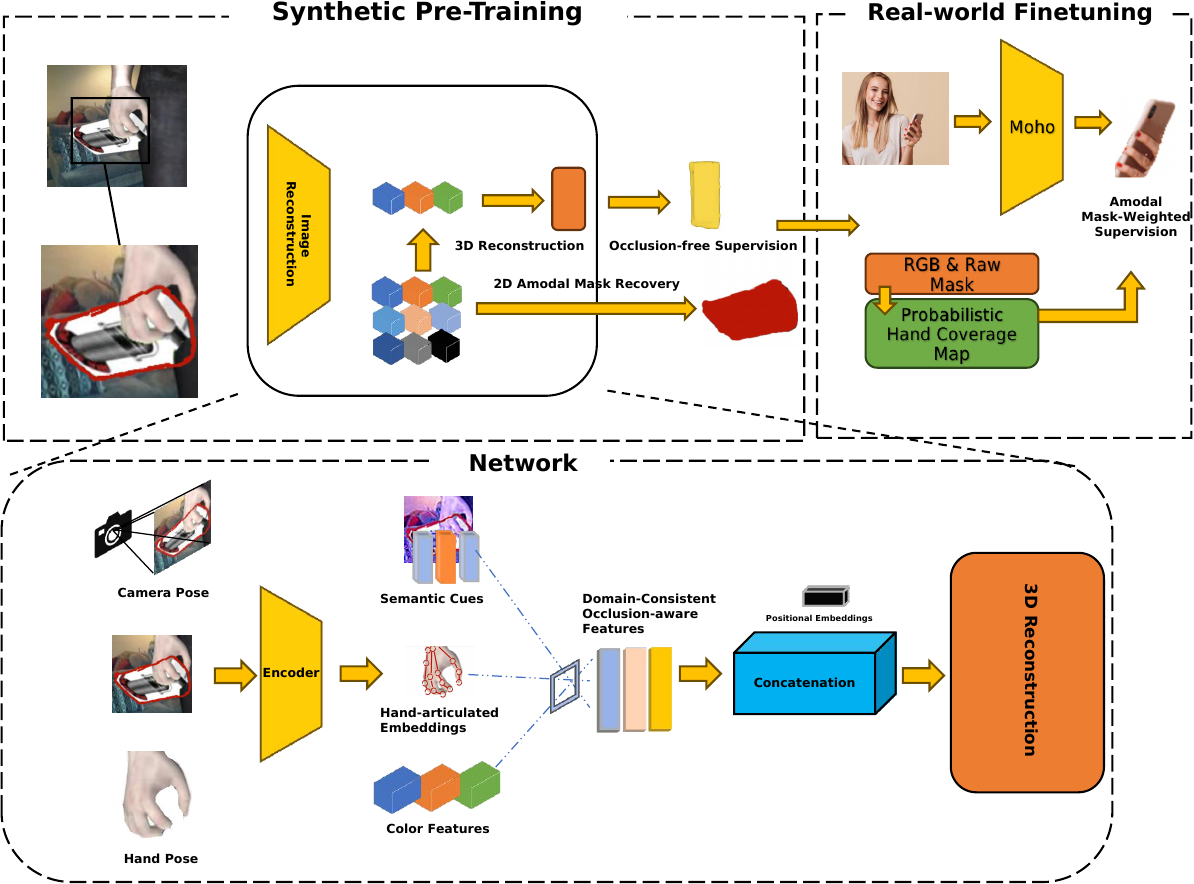}
    \caption{Overview of the proposed second 3D reconstruction module
$f_{3D}$. This method reconstructs a mesh of occluded objects from a single RGB image}
    \label{fig:fig2}
\end{figure}

where $\alpha \in[0,1]$ represents the transparency level of the visual embedding, $\mathbf{x}^v$ represents the merged embedding.

\subsection{Visual Embedding For Occluded Objects}
For the second visual encoder to provide the visual embedding (token) $\mathbf{x}_{\mathrm{v2}}$ of the occluded object, we designed the second visual encoder $f_{3D}$ \cite{li2024mgm}, which is composed as follows:

\begin{figure*}[htbp]
    \centering
    \includegraphics[width=\textwidth]{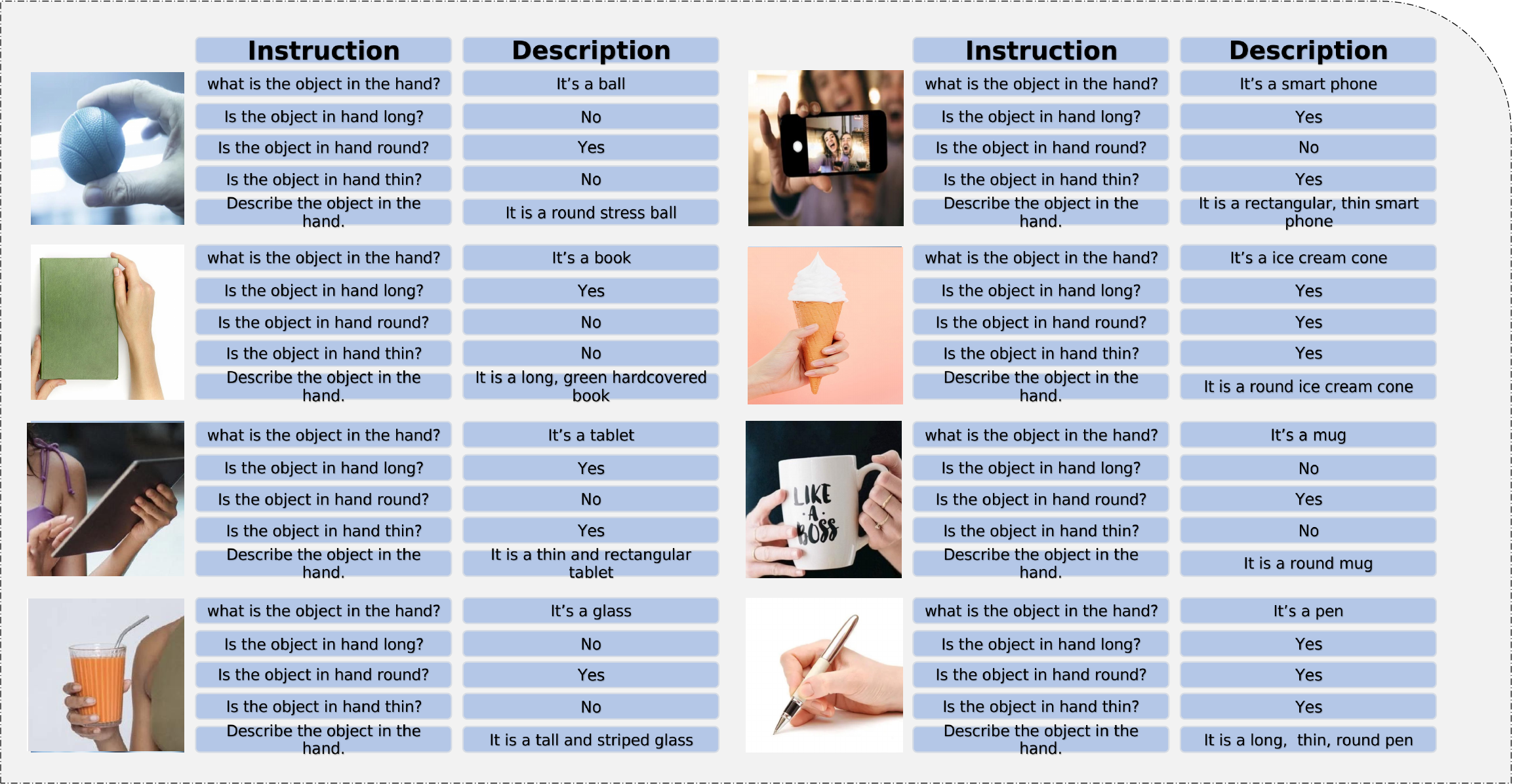}
    \caption{Dataset example. The object is occluded. There are five instructions and five corresponding descriptions.}
    \label{fig:fig3}
\end{figure*}

In the first step, the representation of the semantic cues, hand-articulated features and color features \cite{li2024mgm} of the occluded object are calculated (shown in Figure \ref{fig:fig2}). These representations are merged into a combination of visual features. The calculation is represented as the following:
\begin{equation}
\begin{aligned}
& f_{combined} = f_{s}(f_{cues} + f_{hand} + f_{color}) \\
& \operatorname{SDF}_{\text {object}}(p) = f_{o}(f_{combined}),
\end{aligned}
\end{equation}
where $f_s$ and $f_o$ are the representation accumulation function and SDF decoder, respectively, $p$ represents the 3D point.

In the second step, we apply the calculated SDFs of objects for 3D mesh reconstruction (shown in Figure \ref{fig:fig2}). The computed object $\operatorname{SDF}_{\text {object}}(p)$ already contains the visual representation of the object under occlusion. We reconstruct the 3D mesh $M_{obj}$ of the occluded object and then project it into the 2D RGB space $I_{obj}$. To facilitate the use of this 2D visual representation $I_{obj}$  with large language models, we use the visual embedding of  $\mathbf{x}_{\mathrm{v2}}$ as the extracted embedding of the CLIP model \cite{radford2021learning}. The above calculation is expressed as follows:

\begin{equation}
\begin{aligned}
& M_{obj} = f_{recon}(\operatorname{SDF}_{\text {object}}(p))\\
& I_{obj} = f_{proj}(M_{obj})\\
& \mathbf{x}_{\mathrm{v2}} = f_{CLIP}(I_{obj})\\
\end{aligned}
\end{equation}

\subsection{Test-Time Adaption Based on Self-Supervised Learning.}
To enhance the representation of occluded objects for the multi-modal large language model in the test time, we propose a self-supervised learning strategy with the support of 3D generation module. Specifically, a CLIP model \cite{radford2021learning} is adopted as the reward model and provides feedback for the fine-tuned VLM \cite{li2024mgm}. Given each test sample, with the support of 3D generation module \cite{Zhang_2024_CVPR}, the VLM \cite{li2024mgm} is forced to maximize the CLIP \cite{radford2021learning} reward between the input and sampled results from the fine-tuned VLM \cite{li2024mgm} output distribution.

The self-supervised training is conducted in the reinforcement learning with rewards. In details, the reward is represented as the following:

\begin{equation}
    \mathcal{R}(\boldsymbol{t}, \boldsymbol{v})=\operatorname{CLIP}-\mathrm{S}(\boldsymbol{t}, \boldsymbol{v})-\mathbb{E}_{\boldsymbol{t} \sim P}[\operatorname{CLIP}-\mathrm{S}(\boldsymbol{t}, \boldsymbol{v})]
\end{equation}

Where $\operatorname{CLIP}-\mathrm{S}(\boldsymbol{t}, \boldsymbol{v})$ is the self-supervised clip-score on the base of contrastive learning \cite{radford2021learning}, $\mathbb{E}_{\boldsymbol{t} \sim P}[\operatorname{CLIP}-\mathrm{S}(\boldsymbol{t}, \boldsymbol{v})$ is the corresponding expectation. Where $v$ is the image and $t$ is the corresponding text.

\subsection{Multi-stage Leaning Strategy.}
At the first stage, the VLM \cite{li2024mgm} is fine-tuned on the training dataset \cite{Zhang_2024_CVPR} to perform five specific description tasks (Figure \ref{fig:fig3}). At the second stage, the proposed 3D generation module is trained on the training dataset \cite{Zhang_2024_CVPR} for 3D reconstruction from a single image. At the third stage, to enhance the representation of the occluded objects, the proposed test-time self-supervised adaption strategy is conducted to force the VLM \cite{li2024mgm} in the combination with the 3D generation module \cite{Zhang_2024_CVPR}.

%% file: sec/4_dataset.tex
\section{Dataset}
We use a large-scale dataset SOMVideo \cite{Zhang_2024_CVPR} containing occluded objects to train the proposed multi-modal large language model to understand them.

\subsection{Dataset Overview}
This dataset SOMVideo \cite{Zhang_2024_CVPR} consists of a total of $141,550$ scenes with each hand-object scene captured by $10$ different views. Each corresponding occlusion-free video clip for supervision is also captured from the same $10$ view angles. It also contains $141,550 \times 10 \times 5$ image-text pairs. This dataset was released to describe occluded objects, and to the best of our knowledge, it is for text descriptions of occluded objects. Besides, we manually calculate the occlusions that about a quarter of the objects are occluded on average,   

It is important to note that the annotations(text description) of each sample are manually checked. Furthermore, we apply the proposed dataset in the instruction tuning(fine-tuned) stage. All input images are resized to $224 \times 224$. (Shown in Figure \ref{fig:fig3}). 




%% file: sec/5_Experiment.tex
\section{Experiments and Results}
\subsection{Experiments on GPT4o \cite{gpt4o}}
We first evaluate the performance of GPT4o \cite{gpt4o} on the testing portion of the proposed dataset. Four instructions are applied to test each sample in the testing dataset. And the accuracy is demonstrated in the Table \ref{tab1}. As Table \ref{tab1} shows, the accuracy of the GPT4o \cite{gpt4o} is relatively low. In detail, the accuracy for the instruction $1$(What's the object in the hand?) is $0.1306$, the accuracy for the instruction $2$(Is the object in the hand round?) is $0.6910$, the accuracy for the instruction $3$(Is the object in the hand long?) is $0.6521$, the accuracy for the instruction $4$(Is the object in the hand thin?) is $0.5839$. It demonstrates that GPT4o \cite{gpt4o} cannot achieve satisfactory results for the occluded objects. 

\subsection{Experiments on Mini-Gemini \cite{li2024mgm}}
Then, we fine-tuned one epoch for Mini-Gemini \cite{li2024mgm} using the training set of SOMVideo \cite{Zhang_2024_CVPR}. The hyper-parameter settings for fine-tuning Mini-Gemini \cite{li2024mgm} are set as the following: the batch size is $16$; The learning rate is $0.00002$; The weight attenuation coefficient is $0$. As Table \ref{tab2} shows, in comparison with GPT4o \cite{gpt4o}, the accuracy is higher for instruction $1$, the accuracy is a little higher for instruction $2$, instruction $3$ and instruction $4$. The visual encoder of the proposed Mini-Gemini \cite{li2024mgm} is the common clip encoder\cite{radford2021learning}. (Shown in Figure \ref{fig:fig1}). It demonstrates that fine-tuning on a classical multi-modal large language model \cite{li2024mgm} with a single clip encoder \cite{radford2021learning} improves the accuracy of the instructions from $0.1306$ to $0.4981$. However, the accuracy of $0.4981$ is still not satisfactory.       


\subsection{Experiments on the Proposed 3D Reconstruction Module \cite{Zhang_2024_CVPR}} 
We next explore the capability of the 3D reconstruction module \cite{Zhang_2024_CVPR} for the test description of the occluded objects. At the stage $1$, we train the 3D reconstruction module \cite{Zhang_2024_CVPR} for the task of 3D reconstruction from a single image. At stage $2$, we render the occluded object mesh from the 3D reconstruction module and then project it to 2D RGB space. The rendered RGB image is then described using the fine-tuned VLM \cite{li2024mgm} for each test image.

In the testing phase, we calculate the accuracy of the occluded objects given a single image of the occluded objects. As Table \ref{tab2} demonstrates, in comparison with the fine-tuned VLM \cite{li2024mgm}, the accuracy of the instruction $1$ for falling testing samples \cite{li2024mgm} is $0.1692$. In detail, there are $6258$ occluded samples in the testing set \cite{Zhang_2024_CVPR}, the fine-tuned VLM \cite{li2024mgm} achieves $4366$ correct prediction for the object category classification. Then, the 3D reconstruction module \cite{Zhang_2024_CVPR} achieves $1128$ correct prediction for the left $1892$ falling object samples.

\begin{table}[htbp]
    \centering
    \captionsetup{font={small, bf}} 
    \caption{Experimental results of GPT4o and Mini-Gemini}
    \begin{tabular}{ccc} 
        \toprule[\heavyrulewidth] 
        Model & GPT4o(Zero-shot) & Mini-Gemini \\
        \midrule[\lightrulewidth] 
        Instruction 1 & 0.1306 & 0.4981 \\
        Instruction 2 & 0.6910 & 0.7284 \\
        Instruction 3 & 0.6521 & 0.7325 \\
        Instruction 4 & 0.5839 & 0.7139 \\
        \bottomrule[\heavyrulewidth] 
    \end{tabular}
    \label{tab1}
\end{table}

\begin{table}[htbp]
    \centering
    \captionsetup{font={small, bf}} 
    \caption{Accuracy of classification (Instruction 1) for the 3D reconstruction module among falling samples from fine-tuned VLM \cite{Zhang_2024_CVPR}}
    \begin{tabular}{ccc} 
        \toprule[\heavyrulewidth] 
        Encoder & Task & Accuracy\\
        \midrule[\lightrulewidth] 
        3D Reconstruction \cite{Zhang_2024_CVPR} & Instruction 1 & +0.1692 \\
        \bottomrule[\heavyrulewidth] 
    \end{tabular}
    \label{tab2}
\end{table}

\subsection{Future Experiments}
As the above results demonstrated, the proposed 3D reconstruction module \cite{Zhang_2024_CVPR} is promising for facilitating the understanding of the occluded objects. We plan to further explore this capability in subsequent experiments.

Firstly, the 3D reconstruction module \cite{Zhang_2024_CVPR} continues to be fine-tuned for the task of the instruction $2$, instruction $3$ and instruction $4$. Secondly, the 3D reconstruction module \cite{Zhang_2024_CVPR} is merged with the Vision-Language Model(VLM) \cite{li2024mgm} in a self-supervised learning framework. 

